\documentclass{article} 
\usepackage{colm2024_conference}

\usepackage[utf8]{inputenc} 
\usepackage[T1]{fontenc}    
\usepackage{hyperref}       
\usepackage{url}            
\usepackage{booktabs}       
\usepackage{amsfonts}       
\usepackage{nicefrac}       
\usepackage[pdftex]{graphicx}
\usepackage{microtype}      
\usepackage{xcolor}         
\usepackage{graphicx}
\usepackage{natbib}
\usepackage{ulem}
\usepackage{bbm}
\usepackage{soul}
\usepackage{subcaption}
\usepackage{amsmath} 
\usepackage{algorithm}
\usepackage{algpseudocode}
\usepackage{caption}
\usepackage{amsthm}
\usepackage{multirow}
\usepackage{colortbl}  
\usepackage{lipsum} 
\usepackage{wrapfig}
\usepackage{tcolorbox}
\usepackage{tabularx}
\theoremstyle{plain}

\theoremstyle{definition}

\theoremstyle{remark}

\DeclareMathOperator*{\argmax}{arg\,max}

\definecolor{c1}{RGB}{255,200,200}
\definecolor{c2}{RGB}{150,200,255} 

\newcommand{\fixchen}[1]{\footnote{\textcolor{red}{\textbf{FIX-CHENG!!!} #1}}}

\newcommand{\fixme}[1]{\footnote{\textbf{FIXME!!!} #1}}
\newcommand{\OurMODEL}{\textsc{Temple-MQA}}
\newcommand{\OurDataset}{\textsc{TKeMqa}}
\newcommand{\eat}[1]{}
\newcommand{\warn}[1]{\textcolor{red}{#1}}
\newcommand{\asif}[1]{\textcolor{blue}{#1}}
\DeclareUnicodeCharacter{0301}{\hspace{-1ex}\'{ }}
\newcommand{\di}[1]{\textcolor{green}{#1}}

\title{Multi-hop Question Answering under Temporal Knowledge Editing} 

\author{Keyuan Cheng\thanks{The first three authors contributed equally to this work.}$^{*,1,2,3}$, Gang Lin$^{*,1,2,3}$, Haoyang Fei$^{*,1,2,3}$, Yuxuan zhai$^{3}$, \\  
  \textbf{Lu Yu$^{5}$, Muhammad Asif Ali$^{1,2}$, Lijie Hu\thanks{Correspondence to Lijie Hu \{lijie.hu@kaust.edu.sa\} and Di Wang \{di.wang@kaust.edu.sa\}.}$^{\dagger,1,2,4} $, and Di Wang$^{\dagger,1,2,4}$}\\
  $^1$Provable Responsible AI and Data Analytics (PRADA) Lab\\
  $^2$King Abdullah University of Science and Technology \\
  $^3$South China University of Technology \quad $^4$SDAIA-KAUST AI \quad $^5$Ant Group 
}

\colmfinalcopy 
\begin{document}

\maketitle

\begin{abstract}
Multi-hop question answering (MQA) under knowledge editing (KE) has garnered significant attention in the era of large language models. However, existing models for MQA under KE exhibit poor performance when dealing with questions containing explicit temporal contexts. To address this limitation, we propose a novel framework, 
namely~\underline{\textbf{TEMP}}oral know\underline{\textbf{LE}}dge augmented \underline{\textbf{M}}ulti-hop \underline{\textbf{Q}}uestion \underline{\textbf{A}}nswering (\OurMODEL{}). Unlike previous methods, \OurMODEL{} first constructs a time-aware graph (TAG) 
to store edit knowledge in a structured manner. Then, through our proposed inference path, structural retrieval, and joint reasoning stages, \OurMODEL{} effectively discerns temporal contexts within the question query. Experiments on benchmark datasets demonstrate that~\OurMODEL{} significantly outperforms baseline models. Additionally, we contribute a new dataset, namely \OurDataset{}, which serves as the inaugural benchmark tailored specifically for MQA with temporal scopes. 
\end{abstract}

\vspace{-0.2in}
\section{Introduction}
\label{sec:intro}
\vspace{-0.1in}
Large Language Models (LLMs) have garnered widespread attention owing to their remarkable capacity for knowledge comprehension, enabling tailored solutions across various applications \citep{Zhao2023ASO,Huang2022TowardsRI}. However, the presence of outdated knowledge presents a significant challenge, impeding the ability of LLMs to provide accurate responses regarding recent events and facts -- a phenomenon known as hallucinations, wherein LLMs tend to fabricate plausible yet incorrect responses about unknown facts \citep{Hong2023FaithfulQA}. Thus, ensuring the timely updating of LLMs with the latest information is of paramount importance. However, as the most direct approach for updating information, editing LLMs by re-training from scratch is practically infeasible, as it requires huge computational resources and substantial investments. 
In response, \textit{Knowledge Editing} (KE) has emerged as a focal point, which aims to precisely modify or update the knowledge in LLMs without requiring model retraining. This topic has garnered considerable attention in recent years \citep{2023_KE_Survey,Zhang2024ACS}. 

Multi-hop question answering (MQA), on the other hand, aims to tackle complex inquiries that necessitate multiple reasoning steps. In the context of MQA, KE introduces the concept of \textit{``ripple effects''}, wherein a single edit may trigger a cascade of subsequent knowledge edits or updates \citep{Cohen2023EvaluatingTR}. For instance, if we update the knowledge about the U.S. president from \textit{Trump} to \textit{Biden}, correspondingly knowledge for the question: ``Who is the wife of the U.S. president?'' should also be updated \citep{Cohen2023EvaluatingTR}. There are two prominent lines of work in this area: parameter-based editing and memory-based editing. Parameter-based editing methods update the knowledge by directly modifying the parameters of the model \citep{meng2022locating,meng2022mass}, while memory-based methods employ an explicit memory to store information about the facts to be modified \citep{Mitchell2022MemoryBasedME,zhong2023mquake,gu2023pokemqa}. In most cases, the memory-based approaches 
outperform the parameter-based methods \citep{gu2023pokemqa,zhong2023mquake}

\begin{figure}[t]
    \centering
    \includegraphics[width=0.85\linewidth]{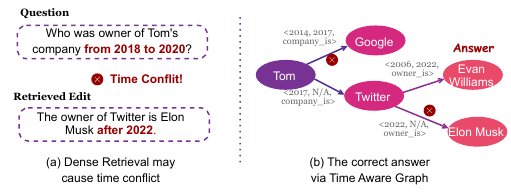}
    \caption{\textbf{Limitations of dense retrieval to handle temporal context.} 
    (a) There is a temporal constraint in question. However, the dense retrieval mechanism fails to accurately capture this temporal information, thus yielding an erroneous retrieval.
    (b) Our time-aware graph stores edits as a time-conscious knowledge structure to help distinguish temporal context and improve retrieval accuracy.}
    \vspace{-3.7ex}
    \label{fig:challenge}
\end{figure}

To address ripple effects, memory-based methods adopt a plan-and-solve paradigm \citep{Khot2022DecomposedPA, Wang2023PlanandSolvePI}, wherein LLMs are prompted to decompose multi-hop questions into sub-questions, followed by iteratively answering each sub-question. These methods employ dense retrieval to identify relevant edits for each sub-question by comparing semantic similarities \citep{Karpukhin2020DensePR}. However, we observed that these approaches exhibit a key limitation, i.e., they perform poorly on questions with explicit temporal contexts \citep{Yin2023HistoryMT}. This inability to cater to temporal context is attributed to the dense retrieval employed by the memory-based methods, which primarily store the edit retrieval information in an unstructured format. This is illustrated in Figure~\ref{fig:challenge} (a), where the dense retrieval retrieves an irrelevant fact (from the year 2022) for the question {\em ``Who was the owner of Tom's company in 2020?''}.

To address the aforementioned limitations, we propose a method called~\OurMODEL{}: \underline{\textbf{TEMP}}oral know\underline{\textbf{LE}}dge augmented \underline{\textbf{M}}ulti-hop \underline{\textbf{Q}}uestion \underline{\textbf{A}}nswering. \OurMODEL{} first constructs a time-aware graph (TAG) to store edit information in a structured format to effectively preserve context-specific temporal information in the best possible way (Figure~\ref{fig:challenge} (b)). To enhance retrieval performance across semantically related edits, \OurMODEL{}: (i) leveraging data augmentation techniques to capture aliases for entity names, aiding in entity disambiguation, and (ii) employing context-dependent concepts to explicitly filter edits based on contextual similarity. To tackle multi-hop questions, \OurMODEL{} utilizes pre-trained LLMs to devise an inference path and conducts step-by-step joint reasoning, leveraging both LLMs and TAG to derive the final response. Experimental evaluation using benchmark data sets 
shows that~\OurMODEL{} outperforms the existing state-of-the-art approaches on MQA by a significant margin. We summarize the key contributions of this work as follows. 

\begin{itemize}
    \item We propose~\OurMODEL{}, the first method capable of achieving high accuracy in MQA under a substantial volume of temporal knowledge edits without forgetting historical knowledge. 
    \item Unlike previous approaches, \OurMODEL{} constructs a TAG to store structured information through our devised methods. Additionally, we propose a novel planning procedure and a joint reasoning approach for the inference path, alongside the development of a unique structural retrieval procedure tailored for knowledge retrieval, with potential applicability to other problems. 
    \item Extensive experimental results on two benchmark datasets show that \OurMODEL{} outperforms the previous seven baselines via different metrics for MQA under massive edits. Furthermore, we develop a new dataset, namely \OurDataset{}, which serves as the first benchmark on MQA with temporal scopes. 
\end{itemize}

\section{Related Work}
\label{sec:RW}
\vspace{-0.1in}
\paragraph{Parameter-based Editing.} Parameter-editing approaches aim to update a model by incorporating information about updated data or knowledge while ensuring minimal changes to predictions on other data points. These approaches can be categorized into fine-tuning, locating and editing, and meta-learning methods. Fine-tuning methods utilize new knowledge to fine-tune the model parameters while at the same time combating catastrophic forgetting \citep{Chen2020RecallAL, Zhu2020ModifyingMI}. Locate and edit approaches treat the layers of a feed-forward network as primary knowledge storage units and update their parameters to edit knowledge. Examples include ROME \citep{meng2022locating} and its extended version, MEMIT \citep{meng2022mass}, which targets a large number of edits. \citet{Huang2024SeeTU} investigated the generalization aspects of knowledge edits, while \citet{Li2023PMETPM} proposed PMET for precise updating of FFN weights. \citet{Hu2024WilKEWK} proposed a wise-layer KE method to facilitate lifelong editing. METO \citep{Yin2023HistoryMT} employs time information as an optimization goal to learn new knowledge without forgetting historical information. Meta-learning approaches treat the editing task as a machine learning challenge, with examples including hyper-networks trained with constrained optimization for fact-only modification \citep{DeCao2021EditingFK}, belief-graph by \citet{Hase2021DoLM}, and 
context-aware meta-learned loss scaling by \citet{Hu2023MetaLearningOA}. However, these methods often underperform for MQA under KE, primarily because the update in the model parameters is more effective for single-hop settings and hard to adapt for multi-hop questions requiring complex reasoning. In contrast, \OurMODEL{} stores the edits in an edit memory alongside a reasoning link, enabling the decomposition and step-by-step solution of multi-hop questions.

{\bf Memory-based Editing.} These techniques store edits in explicit memory and utilize retrieval-augmented methods to reason over relevant edits and modulate end predictions of the model. For instance, SERAC \citep{Mitchell2022MemoryBasedME} introduces semi-parametric editing coupled with a retrieval-augmented counterfactual model. GRACE \citep{Hartvigsen2022AgingWG} embeds additional adapters within LLMs for editing, using vector matching to locate and modify edited knowledge. IKE \citep{Zheng2023CanWE} employs in-context learning based on demonstration storage to edit the model's knowledge. MeLLo \citep{zhong2023mquake} is a simple memory-based
approach that stores all edited facts externally and prompts the language model during inference. Additionally, \citet{zhong2023mquake} also introduces the \textsc{MQuAKE} benchmark for evaluating the MQA performance of their model. Recently, PokeMQA \citep{gu2023pokemqa} proposes a two-stage process of decoupling self-checking and sub-question decomposition to enhance retrieval performance. DeepEdit \citep{Wang2024DeepEditKE} utilizes a depth-first constrained decoding method to edit knowledge for MQA. Note that, unlike existing solutions, \OurMODEL{} does not require sub-problem decomposition but generates an inference path directly, which is more straightforward and can directly generate a complete plan. Moreover, the structural retrieval method proposed by our \OurMODEL{} does not require fine-tuning compared to PokeMQA, making it applicable to a broader range of scenarios, including the processing of temporal information.


\eat{\paragraph{Reasoning over Knowledge Graph.} 
\warn{Cheng: Plz fill in details if we need to add this R/W..!} Yes, sir. Probably not needed. After you finish writing sections 3 and 4, we can discuss them together.}

\vspace{-0.1in}
\section{Preliminaries}
\label{sec:prelimnaries}
\vspace{-0.1in} 
{\bf Notations.}
We represent the set of facts as  $\mathcal{D} = 
\{(s,r,o)\} \subseteq \mathcal{E} \times \mathcal{R} \times \mathcal{E}$, where $\mathcal{E}$, $\mathcal{R}$ denote the set of entities and relations respectively. A tuple $(s,r,o) \in \mathcal{D}$ represents a fact, with subject entity $s$ and object entity $o$ having relation $r$. For the temporal facts, we add time information to $\mathcal{D}$, i.e., $\mathcal{D}_t = \{(s,r,o,\tau_{s}, \tau_{e})\} \subseteq \mathcal{E} \times \mathcal{R} \times \mathcal{E} \times \mathcal{T} \times \mathcal{T}$, where $\mathcal{T}$ represents the timestamps.
A tuple $(s,r,o,\tau_{s}, \tau_{e}) \in \mathcal{D}_{t}$ represents a temporal fact, with subject entity $s$, object entity $o$, having relation $r$ with temporal scope from $\tau_{s}$ to $\tau_{e}$. We use $\mathcal{G}_t$ to represent the time aware graph constructed using $\mathcal{D}_t$ (Section~\ref{sec:TAG}).

\eat{We use $\mathcal{G}$ and $\mathcal{G}_t$ to represent the graphs 
constructed using $\mathcal{D}$ and $\mathcal{D}_t$ respectively.}

\eat{In this section, we provide definitions of the core concepts discussed in this paper, 
i.e., KE, MQA, MQA under KE and \warn{....??}} 
\eat{arguments about MQA under KE as well as the definition of temporal context and formulation of our method.}

\vspace{-0.1in}
\subsection{Knowledge Editing and MQA}
\label{sec:MQA_KE}
\vspace{-0.1in} 
\eat{\paragraph{KG.} 
Knowledge graph (KGs) stores structured knowledge as a graph structure, where a node represents an entity and the relation between them is edge \citep{Ji2020ASO}. Time-conscious knowledge graph to explicitly store temporal information along the graph edges \citep{Jiang2016TowardsTK}.} 

\textbf{Knowledge Editing (KE).}
\eat{Similar to existing research on knowledge editing \citep{zhong2023mquake, Wei2023AssessingKE}, we use the relational triplet $(s,r,o)$ to represent a fact, where $s$ represents the subject entity, and $o$ represents the object entity of the relation $r$.}
A KE request can be represented as $\mathcal{F}=\{f_1,f_2,\cdots f_n\}$, which contains a set of fact edits. Each edit $f_i \in \mathcal{F}$ represents an individual knowledge editing operation, denoted by $f_i=(s,r,o) \rightarrow (s,r,o^*)$, indicating that the object of the subject $s$ with relation $r$ is updated from $o$ to $o^*$.

\eat{Here, each step of the chain $(s_i,r_i,o_i)$ associated to one knowledge. The subject and object are chained together, that is, the $o_i$ from one fact identical to the $s_{i+1}$ of the next fact. $(s_i,r_i,o_i^*)$}

{\bf MQA under KE.} 
A multi-hop question $Q$ requires multiple reasoning steps 
to derive the final answer. The reasoning steps could be represented as a \textit{chain of facts} $\langle(s_1,r_1,o_1),\cdots,(s_n,r_n,o_n)\rangle$, where each step (hop) in the chain $(s_i,r_i,o_i)$ associated with an individual fact. The subject and object are chained together, i.e., the object $(o_i)$ from the preceding fact becomes the subject $(s_{i+1})$ for the subsequent fact. \eat{We use $P = s_1 \xrightarrow{r_1} o_1 \xrightarrow{r_2} \cdots \xrightarrow{r_n} o_n$ to represent the inference path of this chain of facts in $Q$.}

If one of the steps $(s_i,r_i,o_i)$ in  $Q$ is associated with a fact edit $f_i \in \mathcal{F}$, it causes ripple effects, i.e., all subsequent facts have to be updated in order to derive the final answer to the question. Mathematically, a chain of facts in $Q$ coupled with corresponding fact edits may 
be represented as: $\langle(s_1,r_1,o_1),
\cdots,(s_i,r_i,o_i^*),
\cdots,(s_n^*,r_n,o_n^*)
\rangle$. Note that $Q$ may initiate multiple fact edits in $\mathcal{F}$. The end goal of MQA under KE is to derive the final answer $o_n^*$ for $Q$ after considering all associated edits in $\mathcal{F}$.

\vspace{-0.1in}
\subsection{Temporal Scope}
\label{sec:T-MQA}
\vspace{-0.1in} 
{\bf Temporal KE.} In KE under temporal scope,  we use $\mathcal{F}_t = \{f_1, \cdots, f_n\}$ to represent the temporal fact edits, with a single temporal fact edit denoted by $f_i=(s,r,o,\tau_s,\tau_e)\rightarrow(s,r,o^*,\tau_s^*,\tau_e^*)$.
It means that $(s,r,o)$ is valid from $\tau_s$ until $\tau_e$, and after that $(s,r,o^*)$ is valid from $\tau_s^*$ to $\tau_e^*$.

{\bf Temporal MQA under KE.} Similar to MQA under KE, here, each reasoning step will be represented as a temporal fact. A chain of facts in $Q$ coupled with corresponding temporal fact: 
$\langle(s_1,r_1,o_1, \tau_{1, s},\tau_{1, e}),
\cdots,(s_i,r_i,o_i^*, \tau^*_{1, s},\tau^*_{1, e}),
\cdots, (s_n^*,r_n,o_n^*,\tau^*_{n, s},\tau^*_{n, e})
\rangle$.

\begin{figure}[t]
    \centering
    \includegraphics[width=\linewidth]{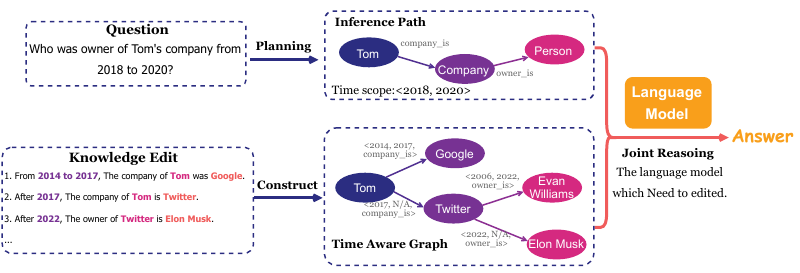}  \caption{\textbf{Overview of~\OurMODEL{}}. Given a multi-hop question,~\OurMODEL{} employs LLMs to strategize an inference path for the question. Then, it leverages LLMs and the TAG ($\mathcal{G}_{t}$) for joint reasoning on the inference path.}
    \label{fig:Vs}
    \vspace{-12pt}
\end{figure}

\vspace{-0.1in}
\section{\OurMODEL{}}
\label{sec:proposed}
\vspace{-0.1in} 
\OurMODEL{} is a generalized memory-based editing method. It is capable of MQA under knowledge editing (including knowledge correction and updating, etc.) without forgetting historical knowledge. As shown in Figure~\ref{fig:Vs}, the core components 
of~\OurMODEL{} include pre-existing LLMs and a Time Aware Graph (TAG).
The workflow of~\OurMODEL{} is summarized as follows: 
(i) Constructing a TAG for the given edits;
(ii) Utilizing LLMs to devise an inference path for each multi-hop question;
(iii) Finally, employing the LLMs along with TAG via our proposed structural retrieval mechanism for joint reasoning based on the inference path. We will discuss each component in detail.

\vspace{-0.1in}
\subsection{Time-Aware Graph Construction}
\label{sec:TAG}
\vspace{-0.1in} 
The objective of the graph construction process is to design a mechanism to store knowledge edits in a structured format, facilitating effective and efficient retrieval. It is worth noting that this approach markedly differs from existing approaches, which predominantly store knowledge edits in an unstructured format \citep{zhong2023mquake}. This distinction proves crucial in overcoming the challenge of handling temporal contexts for KE. 

Historical and current facts are typically stored in unstructured text form as $f_{\text{old}}$ and $f_{\text{cur}}$. The graph construction process initially transforms them into structured form $\mathcal{F}_t$ (detailed in Section \ref{sec:T-MQA}). Then it utilizes these structured temporal edits $\mathcal{F}_t$ as input and generates a graph $\mathcal{G}_{t}$ as output. The process flow is elucidated as follows.

{\bf Converting unstructured facts to a structured form.}
First, we employ a pre-trained LLM to convert all unstructured facts $f_x$ (include $f_{\text{old}}$ and $f_{\text{cur}}$) into a structured form to explicate the information content across different dimensions. To achieve this, we design an in-context learning prompt (see Appendix \ref{sec:prompt} for details) that prompts an LLM to process the unstructured fact.
\begin{equation}
 s,r,o,\tau_s,\tau_e,c(o) = \text{LLM}(\text{P}_{convert}(f_{\text{x}})),
\label{eq:convert1}
\end{equation}
Where $f_\text{x}$ denotes the unstructured fact, LLM yields a structured output $(s,r,o,\tau_s,\tau_e,c(o))$. Here, $c(o)$ represents the concept associated with the object entity $o$~\citep{2020_fgnet,FGERH}, and $\text{P}_{convert}$ is the in-context learning prompt. We use GPT-3.5-turbo-instruct as the pre-trained LLM. Finally, for all pairs of historical and current facts ($f_{\text{old}}$, $f_{\text{cur}}$), via our above procedure, we can get the set of structured temporal fact edits $\mathcal{F}_t = \{ f_x \rightarrow (s,r,o,\tau_s,\tau_e) \}$.

It is notable that compared to the original temporal fact, here we also include the concept information for the object, $c(o)$, with the fact structure. It provides multi-faceted benefits, including (i) disambiguation of the entities to their appropriate concepts based on the context and (ii) improved performance for edit retrieval by offering a wide range of fine-grained concepts~\citep{2012_FIGER}. For instance, in different contexts, the entity ``Washington'' can refer to a city name or a person's name, and without including the concept $c(o)$, explicitly storing this information poses a challenge to distinguishing between them.

{\bf Data augmentation.} 
The purpose of data augmentation is to capture different possible lexical variations for the same entity. This step is pivotal for improving retrieval performance, as it assists in filtering out edits with subjects different from the query subject. This eventually helps reduce the search space by filtering numerous irrelevant edits. We utilize SPARQL (detailed in Appendix \ref{sec:SPARQL}) to get all possible
aliases of the subject entity $s$ from Wikidata, denoted as $\text{A}(s)=\{s_0,s_1,s_2,\cdots\}$, where $s_0=s$ and $s_i$ is the $i$-th alias for $s$. Note that the current formulation of \OurMODEL{} favors data augmentation on the subject dimension.

{\bf Graph construction.}
Finally, we utilize the structured information about historical and current knowledge to construct the TAG $\mathcal{G}_t$. Specifically, we store the current knowledge as $\{(n_{s}:s_i, e:(r,c(o),\tau_s^*,\tau_e^*), n_{e}: o^*)\}$ for $s_i \in \text{A}(s)$, and the historical knowledge as $\{(n_{s}:s_i,e:(r,c(o),\tau_s,\tau_e), n_{e}: o)\}$ for $s_i \in \text{A}(s)$. Here $n_s$, $e$, and $n_o$ represent the start node, edge, and end node in $\mathcal{G}_t$, respectively.

We can also use the above steps to construct graphs for non-temporal edits with the distinction that we exclude temporal scopes. 
In this scenario, we only store the updated knowledge as $\{(n_{s}:s_i, e: (r,c(o)), n_{e}: o^*)\}$ for $s_i \in \text{A}(s)$. Note, for $\mathcal{G}$, we only store current knowledge.


\subsection{Planning and Reasoning on Inference Path}
In this section, we introduce the planning and reasoning stages 
of~\OurMODEL{}: i)  The planning stage leverages LLMs to generate an inference path $P$ and extract temporal scope for the question $Q$. ii) The reasoning stage performs a step-by-step joint reasoning by utilizing LLMs and the TAG we constructed in Section~\ref{sec:TAG} to solve the question $Q$ using the inference path $P$. We will discuss them in detail. 

\textbf{Stage 1: Inference Path.} 
This stage aims to exploit the instruction-following ability of LLMs to generate a structural inference path $P$ that is helpful for answering $Q$. To achieve this, we design an in-context learning prompt (explained in Appendix~\ref{sec:prompt}) that prompts the LLMs to generate an inference path along with the temporal scopes, as shown below. 
\begin{equation}
   P,\tau_s^Q, \tau_e^Q = \text{LLM}(P_{\text{infer}}(Q)),
   \label{eq:Stage1}
\end{equation}
where $P_{\text{infer}}$ is the in-context prompt for $Q$ used 
as input for LLM, yielding the inference path $P$ and temporal scopes $\tau_s^Q$ and $\tau_e^Q$ as outputs. $P$ is represented as $\langle(s_1, r_1, c(o_1)),\cdots, (c(o_{n-1}), r_n, c(o_{n})\rangle$, where $s_1$ is an entity extracted from question $Q$. 

Note that, unlike previous research, our approach is more practical and efficient, as it computes the inference path in one go without needing to alternate between multiple different and solve phases. The inference path has the concept of entity added among the relation compared to the relation path \citep{Luo2023ReasoningOG}, which helps us in retrieval.

\textbf{Stage 2: Joint Reasoning.} As an iterative process, this stage aims to obtain the final answer for $Q$ by reasoning on $P$. 
 \eat{Consider the $i$-th step with query $(s_i,r_i,c(o_i),\tau_s^Q,\tau_e^Q)$ for $1\leq i \leq n$,  we need to infer the specific entity $o_i$.}
Consider the $i$-th step ($1\leq i \leq n$), we utilize $(s_i,r_i,c(o_i),\tau_s^Q,\tau_e^Q)$ as input to infer the specific entity $o_i$, where $s_i$ is the output of last step $o_{i-1} (i>1)$ or start point $s_1$ in $P$, $r_i$ and $c(o_i)$ originate from $P$.
For this, \OurMODEL{} uses the structural retrieval $\mathbf{R}_\text{struct}$ (Section~\ref{sec:str_ret}) 
to retrieve relevant knowledge from the $\mathcal{G}_t$ for the 
query.
\begin{equation}
   o_{R}^*, \alpha_R = \mathbf{R}_\text{struct}(s_i,r_i,c(o_i),\tau_s^Q,\tau_e^Q),
\label{eq:retrieve}
\end{equation}

\begin{figure}
    \centering
    \includegraphics[width=0.85\linewidth]{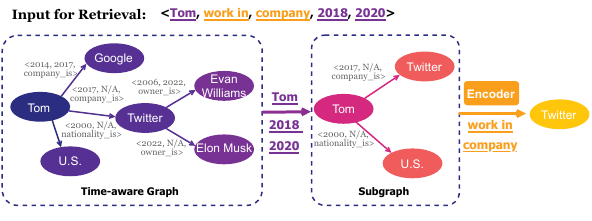}
    \caption{\textbf{The workflow of structural retrieval.} 
    1) The first step is to extract the 1-hop sub-graph where the center point is Tom. We filter out the edits not conforming with the temporal scope of the query\eat{which is too recent or too early}. 
    2) The second step calculates the semantic similarity of the relation and concept between the query and knowledeg in sub-graph.}
    \label{fig:e2}
\vspace{-12pt}
\end{figure}
where $o_{R}^*$ is the possible updated knowledge about $o_i$, 
and $\alpha_R$ denotes the similarity between updated knowledge and query. Note that in our approach, there is no need to use self-checking \citep{zhong2023mquake} to check if the retrieved fact contradicts the generated answer.

Finally, based on the similarity score $\alpha_R$ compared against a
threshold ($\theta$), we decide if we may use retrieved 
knowledge as such or use the language model for response generation: 
\begin{equation}
   o_i=
\begin{cases}
o_{R}^* &  \alpha_R > \theta\\
\mathbf{M}_\text{query}(P_{\text{query}}(s_i,r_i,c(o_i),\tau_s,\tau_e))&  \alpha_R \leq \theta, 
\end{cases}
    \label{eq:retrieveanswer}
\end{equation}
where $\mathbf{M}_{\text{query}}$ is an LLM
, and $P_{\text{query}}$ is the prompt for query {(details are given in Appendix~\ref{sec:prompt})}. We re-iterate the above-mentioned process until we completely exhaust the inference path $P$, yielding $o_n$ as the final answer.

\vspace{-0.1in}
\subsection{Structural Retrieval}
\label{sec:str_ret}
\vspace{-0.1in} 
Given a query $(s_i,r_i,c(o_i),\tau_s^Q,\tau_e^Q)$, this section introduces how to retrieve the knowledge from 
$\mathcal{G}_t$ using structural retrieval $\mathbf{R}_\text{struct}$. The workflow of $\mathbf{R}_\text{struct}$ (See Figure~\ref{fig:e2}) can be represented by the following steps:
(i) Extracting a subgraph, i.e., filtering out the knowledge that does not meet the required subject and temporal scope. 
(ii) Re-ranking, i.e., re-ranking the remaining knowledge based on semantic similarity of relation and concept with query.

\textbf{Step 1: Extracting a Subgraph.} This step extracts a 1-hop subgraph $\mathcal{G}_{sub}$ from $\mathcal{G}_t$ by filtering out relation pairs violating the subject and temporal scopes. Formally, this is represented as:
\begin{equation}
   \mathcal{G}_{sub} = \{(n_s, e, n_e)\in \mathcal{G}_t \mid  (n_s == s_i) \land  
   (e[\tau_s] \geq \tau_s^Q) \land (e[\tau_e] \leq \tau_e^Q) \}, 
    \label{eq:Re1}
\end{equation}
where $n_s$ indicates the subject and ($e[\tau_s],e[\tau_e]$) represent the temporal scope of the fact.

\textbf{Step 2: Re-ranking.} 
This step aims to re-rank the candidate answers and find the one with the highest similarity.
For this, we first use an existing encoder $E$ to encode the relation and concept for the query, as follows:
\begin{equation}
   v_{r}^q=E(r_i); v_{c(o)}^q=E(c(o_i)). 
    \label{eq:Ret1}
\end{equation}
Finally, we re-rank knowledge in $\mathcal{G}_{sub}$ based on the cosine similarity ($sim$) for both relation and fine-grained concept 
between knowledge and query and then select the knowledge with the highest similarity as the retrieved result.
\begin{equation}
 o_R^*,\alpha_R = \argmax_{(n_s, e, n_e) \in \mathcal{G}_{sub}} ({sim(E(e[r]),v_r^q)+sim(E(e[c(o)]),v_{c(o)}^q)}). 
    \label{eq:Ret3}
\end{equation}
Where $o^*_R$ is the retrieved object, and $\alpha_R$ is the corresponding similarity returned by $\mathbf{R}_\text{struct}$, $e[r]$ and $e[c(o)]$ indicate relation and concept of the knowledge.

\vspace{-0.1in}
\section{Experiments}
\label{sec:exp}
\vspace{-0.1in} 
Here we conduct practical evaluations for~\OurMODEL{} compared against different baseline models.  
Additional results, including the ablation study, can be found in Appendix \ref{sec:more}. 
\vspace{-0.05in}
\subsection{Experimental Settings}
\vspace{-0.05in} 
\textbf {Datasets.}
We evaluate~\OurMODEL{} on a blend of publicly available 
benchmarks and self-curated datasets. These include:
\textsc{MQuAKE}~\citep{zhong2023mquake},
\textsc{AToKe}~\citep{Yin2023HistoryMT}, and our newly proposed dataset \OurDataset{}. Detailed descriptions and statistics of the datasets are given in Appendix~\ref{appendix:Dataset} and~\ref{sec:ourdata}. 

\textbf{Baselines.}
We compare the performance of~\OurMODEL{} against a wide range of parameter-based and memory-based KE methods. The parameter-based baselines include: 
{Fine-tuning (FT)}~\citep{Zhu2020ModifyingMI},
{ROME}~\citep{meng2022locating},
{MEMIT}~\citep{meng2022mass}, MEND \citep{Mitchell2021FastME} and
{METO}~\citep{Yin2023HistoryMT}.
The memory-based baselines include: 
{MeLLo}~\citep{zhong2023mquake},
and
{PokeMQA}~\citep{gu2023pokemqa}.
Details are given in Appendix~\ref{Appendix:baseline}.

\textbf{Evaluation metrics.}
Similar to the baseline models, we use five different evaluation metrics: 
(i) Multi-hop Accuracy (M-Acc)~\citep{zhong2023mquake},
(ii) Hop-wise Accuracy (H-Acc)~\citep{gu2023pokemqa},
(iii) Historical Explicit time Question Score (HES)~\citep{Yin2023HistoryMT}, 
(iv) Current Explicit time Question Score (CES)~\citep{Yin2023HistoryMT}, and
(v) Historical Explicit time Question Score (CES-P)~\citep{Yin2023HistoryMT}. For all metrics, a larger value indicates that the method is better. It is notable that we do not test the performance of~\OurMODEL{} with relative and/or implicit temporal metrics because~\OurMODEL{} requires explicit temporal scope 
to distinguish historical knowledge from existing knowledge. Details of evaluation metrics are given in Appendix~\ref{Append:evalm}.

\textbf{Experimental setup.}
To evaluate performance with varying numbers of edits, we follow the setting of PokeMQA \citep{gu2023pokemqa} to conduct stratified sampling \citep{parsons2014stratified} of the dataset based on the number of hops in questions. This allows us to construct edit batches of different sizes while ensuring a relatively consistent proportion of questions with different hop counts within each batch. We inject all the edits within a batch simultaneously \citep{zhong2023mquake}. The scenarios of different batch sizes are denoted as M-edited ($M\in \{1,100,All\}$). We use LLaMa-2-7B~\citep{Touvron2023Llama2O},  Vicuna-7B~\citep{vicuna2023}, GPT-turbo-3.5-instruct and  GPT-J-6B \citep{gpt-j} as the LLMs and utilize \texttt{all-MiniLM-L12-v2}\footnote{\url{https://huggingface.co/sentence-transformers/all-MiniLM-L12-v2}} as the encoder in our method. We conducted each experiment four times and reported the average values in the table. All experiments are performed using PyTorch 2.1.2 and RTX 4090 24GB GPU. 
\vspace{-0.05in}
\subsection{Experimental Results}
\label{sec:main-results}
\vspace{-0.05in} 
\paragraph{\bf Non-temporal MQA under KE.}
We initially test the performance of~\OurMODEL{} for non-temporal MQA, whose results for~\textsc{MQuAKE} are shown in Table~\ref{tab:exp1}. These results clearly show that~\OurMODEL{} consistently outperforms the baseline models across different settings by a large margin. For example, when consider the ~\textsc{MQuAKE-CF-3K} dataset and M-Acc as the evaluation metric, \OurMODEL{} on average improved by 91.2\%, 112.7\% and 101.4\% compared to Mello for \{$1, 100, All$\}-edited respectively across three LLMs, and by 42.6\%, 32.4\%, and 28.6\% compared to PokeMQA. We attribute such drastic performance improvement to the following factors:
(a) The inference path employed by~\OurMODEL{} is more reliable, which helps the model to boost the end performance significantly; 
(b) The structural retrieval pipeline (i.e., TAG) designed for~\OurMODEL{} is robust to reduce the number of false positives significantly. We provide an analysis of these factors in Section~\ref{sec:ablation}. 

Compared \OurMODEL{} with different LLMs, we can see it exhibits superior performance on GPT-3.5-turbo-instruct followed by Vicuna-7B. This is due to the fact that GPT-3.5-turbo-instruct has the strongest reasoning capabilities \citep{vicuna2023}, enabling our model to generate more reliable inference paths. Another noteworthy observation in Table~\ref{tab:exp1} is that parameter-based approaches perform poorly compared to memory-based ones, which is consistent with the results in \citep{zhong2023mquake}. 
\begin{table*}[t]
\centering
\resizebox{0.9\linewidth}{!}{
\begin{tabular}{cccccccccccc}
\toprule[1.0pt]
\multirow{3}{*}{\textbf{Method}} & \multicolumn{6}{c}{\textbf{\textsc{MQuAKE-CF-3K}}} & \multicolumn{4}{c}{\textbf{\textsc{MQuAKE-T}}} \\ \cmidrule{2-11} 
& \multicolumn{2}{c}{1-edited}    & \multicolumn{2}{c}{100-edited}     & \multicolumn{2}{c}{All-edited}  & \multicolumn{2}{c}{1-edited}   & \multicolumn{2}{c}{All-edited}       \\ \cmidrule{2-11} 
& M-Acc  &H-Acc    & M-Acc  &H-Acc   & M-Acc  &H-Acc   & M-Acc  &H-Acc   & M-Acc  &H-Acc 

\\ \hline

\multicolumn{11}{c}{\cellcolor{gray!25}\scshape LLaMa-2} 
\\ \hline
FT$^*$        & 28.20  & 7.3   & 2.37  & 0.03 & -    & -   & 56.48 & 33.89  &1.02   &0.37      \\
ROME$^*$    & 13.13  &5.37 & 3.50 &0.03   & 3.63  &0.1 & 24.89 &17.99  & 1.71   &0.32 \\ 
MEMIT$^*$   & 14.97 &6.43  & 9.40 &2.47   & 2.30 &0.37  & 30.89 &23.98  & 25.21  &20.13 \\
MeLLo   & 33.57  &9.9 & 20.00 &10.07   & 17.33 &9.9  & 65.78 & 55.27  & 57.69  &44.55  \\
PoKeMQA$^*$   & \underline{44.13}  &\underline{30.6} & \underline{37.33}  &\underline{27.83}  & \underline{32.83} &\underline{23.87}  & \underline{75.43} &\underline{60.44}  & \underline{74.36}   &\underline{60.22} \\
\OurMODEL{} (Ours)   & \textbf{68.32}   & \textbf{59.46}  & \textbf{48.95}   & \textbf{35.17}  & \textbf{42.20}  & \textbf{27.51}  & \textbf{77.56}  & \textbf{64.87} &\textbf{75.73}   & \textbf{62.30}   \\
\hline
\multicolumn{11}{c}{\cellcolor{gray!25}\scshape Vicuna-7B} 
\\ \hline
MeLLo$^*$    & 30.70  &20.84  & 24.75  & 12.25 & 22.35  &10.18 & 60.72 & 48.55   & 51.55   & 42.97 \\
PoKeMQA$^*$   & \underline{45.83} &\underline{34.83}  &\underline{38.77} &\underline{31.23}  & \underline{31.63} &\underline{25.3}  & \underline{74.57}  &\underline{55.19} & \underline{73.07}  &\underline{55.09}  \\
\OurMODEL{} (Ours)   & \textbf{71.61}   & \textbf{62.75}   & \textbf{56.65}  & \textbf{44.26}   & \textbf{46.60} & \textbf{37.33}  & \textbf{81.77}  & \textbf{69.46}  & \textbf{78.29}  & \textbf{68.15}  \\ \hline
\multicolumn{11}{c}{\cellcolor{gray!25}\scshape GPT-3.5-turbo-instruct} 
\\ \hline
MeLLo$^*$    & 57.43  &28.8 & 40.87   &28.13 & 35.27 &25.3  & \underline{88.12}  &52.84 & 74.57 &53.53   \\
PoKeMQA$^*$    &\underline{67.27}  &\underline{56.37} & \underline{56.00}  &\underline{49.63}  & \underline{48.87}  &\underline{39.77} & 78.16 &\underline{68.09}  &\underline{76.98}  &\underline{67.88}  \\
\OurMODEL{} (Ours)   & \textbf{78.11} & \textbf{63.45}  & \textbf{67.21} &\textbf{55.33}   & \textbf{53.68}  &\textbf{40.05} & \textbf{90.57} &\textbf{81.90}  & \textbf{82.26} &\textbf{74.33} \\

\bottomrule[1.0pt]
\end{tabular} }
\caption{\textbf{Experimental results for~\textsc{MQuAKE}.} 
We \textbf{boldface} the best-performing scores with the second-best \underline{underlined}. The result from the previous paper is marked as *. By default, the same symbols are used in the following tables.}
\vspace{-3.7ex}
\label{tab:exp1}
\end{table*}

\paragraph{Single-hop MQA under Temporal KE.}
We then consider single-hop question answering with temporal knowledge, whose results for \textsc{AToKe} are shown in Table~\ref{tab:exp2}. Compared to  Table~\ref{tab:exp1}, the results show varying behavior, with parameter-based approaches outperforming memory-based methods in some cases, which is contributed by their stronger ability to retain updated knowledge. 
Specifically, we can see that parameter-based approaches (such as ROME and MEMIT) can achieve very good performance for CES and CES-P, indicating that they can effectively inject new knowledge into the model. However, they perform poorly on the HES metric: even after using METO, the average HES is only 30.5\%.  This indicates that these methods forget historical knowledge after editing.

On the other hand, MeLLo and PokeMQA achieved higher HES scores but lower CES scores, indicating they are unable to distinguish historical knowledge and edited knowledge. \OurMODEL{} can take into account both historical and new knowledge. It not only has higher HES than MeLLo and PokeMQA but also maintains high CES. Specifically, on \textsc{AToKe-SE}, \OurMODEL{} achieves HES of 97.25, an improvement of 221\% compared to MEMIT$_{\text{METO}}$. The reason for such a big improvement is that \OurMODEL{} can effectively distinguish different knowledge in the same time chain through TAG. 


\begin{table*}[htbp]
\centering
\resizebox{\linewidth}{!}{
\begin{tabular}{cccccccccccccc}
\toprule[1.0pt]
\multirow{3}{*}{\textbf{Method}} & \multicolumn{6}{c}{\textbf{\OurDataset{}}} & \multicolumn{6}{c}{\textbf{\textsc{TKeMqa-HK}}} \\ \cmidrule{2-13} 
& \multicolumn{2}{c}{1-edited}    & \multicolumn{2}{c}{100-edited}     & \multicolumn{2}{c}{All-edited}  & \multicolumn{2}{c}{1-edited} & \multicolumn{2}{c}{100-edited}  & \multicolumn{2}{c}{All-edited}       \\ \cmidrule{2-13} 
& M-Acc  & HES  & M-Acc  & HES  & M-Acc  & HES & M-Acc  & HES & M-Acc  & HES & M-Acc  & HES 

\\ \hline

\multicolumn{13}{c}{\cellcolor{gray!25}\scshape Vicuna} 
\\ \hline
MeLLo    & \underline{39.02}  & \textbf{36.54} & \underline{31.71}   & \textbf{35.56}  & \underline{25.51}  & \underline{32.21}  & \underline{25.76}   &\underline{62.76}  & 18.53   & \underline{60.26}  & 15.21   & \underline{56.89}    \\
PoKeMQA & 36.65   & 33.12 &  30.15   & 31.90  & 23.02  & 29.56  & 22.13   & 57.83  & \underline{20.35}   & 56.71  & \underline{16.94}   & 52.61  \\
\OurMODEL{} (Ours)    & \textbf{52.18}   & \underline{35.65}  & \textbf{50.41}   & \underline{34.62}  & \textbf{48.13}  & \textbf{34.51}  & \textbf{51.92}   &\textbf{77.51}  & \textbf{50.13}   & \textbf{75.59}  & \textbf{48.20}   & \textbf{75.52}  \\ \hline

\multicolumn{13}{c}{\cellcolor{gray!25}\scshape GPT-3.5-turbo-instruct} 
\\ \hline
MeLLo    & \underline{82.14}   & \underline{37.75}  & \underline{59.17}   & \underline{37.19}  & \underline{47.53} & \textbf{36.57}  & \underline{56.04}   & \underline{73.44}  & \underline{38.22}   & \underline{72.63} & \underline{31.45}   & \underline{70.75}   \\
PoKeMQA   & 70.18   & 27.85  & 53.54   & 24.55  & 43.80  & 20.26  & 50.27   & 67.91  & 33.83   & 64.37  & 27.42   & 60.65 \\
\OurMODEL{} (Ours)   & \textbf{85.53}   & \textbf{41.26}  & \textbf{83.47}  &\textbf{38.75}  & \textbf{80.50}  & \uline{33.11}  & \textbf{85.51}   & \textbf{82.28}  & \textbf{84.12}   & \textbf{82.22}  & \textbf{80.09}   & \textbf{81.61} \\

\bottomrule[1.0pt]
\end{tabular}}
\caption{\textbf{Experimental results on \textsc{TKeMqa}.} 
}
\label{tab:exp3}
\vspace{-12pt}
\end{table*}

\paragraph{MQA under Temporal KE.} We finally consider the performance on \OurDataset{}, which consist of $\{1,2,3,4\}$-hops equation answering with temporal knowledge. The results are shown in Table~\ref{tab:exp3}. We can see that similar to the above results, \OurMODEL{} achieves good performance for M-Acc. However,  \OurMODEL{} is worse than MeLLo for HES in the All-edited setting using GPT. This is because there is no historical knowledge matching the question in the edit memory in such a setting. MeLLo's utilization of a self-checking method effectively validates the retrieved results, contributing to its superiority. 

\eat{We can find that when the edit batch size increases, \OurMODEL{} experiences only a slight performance decrease, while Mello and PokeMQA drop a lot. This is because Mello and PokeMQA match the embeddings of questions and edits for similarity, which easily leads to the loss of temporal information. Since \OurDataset{}-HK contains historical knowledge, the retrieval process of Mello and PokeMQA is prone to be influenced by historical knowledge, leading to incorrect answers. Our method is able to address these limitations by using a carefully crafted inference plan and TAG for edit retrieval to perform the end task in a performance-enhanced fashion. } 

Moreover, in contrast to the previous two datasets, we find that \OurMODEL{} on \OurDataset{} experiences only a slight performance decrease 
when the edit batch size increases.  We further conduct an experiment to study the performance w.r.t. difference edit numbers (see Figure \ref{fig:combined_fig1} in Appendix). It can be observed from Figure \ref{fig:combined_fig1} (a) and Figure \ref{fig:combined_fig1} (b)  that as the number of edits increases, the performance of \OurMODEL{} drops rapidly on the \textsc{MQuAKE-CF-3K}. We find that this is because \textsc{MQuAKE-CF-3K} has unpassable data in the mass-edit setting. Specifically, there are data that are affected by the editing of other data. Under M-edited ($M>1$) settings, if they are in the same batch, unpassable data will be generated (see Appendix Table~\ref{tab:data_conflict} for an example). To verify this, we filter out conflicting data and create a new data \textsc{MQuAKE-CF-3K-Fix}. The result is shown in Figure \ref{fig:combined_fig1} (c). After repairing, we can observe that \OurMODEL{} is more stable than other methods, which illustrates the accuracy of structural retrieval.


\eat{With the increase of edit batch sizes, \OurMODEL{} experiences only a slight performance decrease. However, the performance of MeLLo and PokeMQA dropped a lot at all-edited setting. \warn{It is worth noting that edit used for updating in \OurDataset{} has the same temporal scope, which are applied after 2023. The high performance on M-Acc shows that structured retrieval of \OurMODEL{} is effective without relying on the temporal dimension.}}
\eat{\textbf{MeLLo is reliable and generalized.}}

\eat{MeLLo achieve an average improvement of \warn{fix-me}\% over PokeMQA with M-Acc on \textsc{TKeMqa}. We find that this was due to MeLLo's self-checking. MeLLo uses self-checking to pass the retrieved edits to LLM to determine whether they are relevant, which is very effective when the retrieval accuracy is not high. In Table\ref{tab:exp1} and \ref{tab:exp2}, the retrieval difficulty is lower, so PokeMqa performs better.}
\begin{table*}[t]
\centering
\resizebox{0.75\linewidth}{!}{
\begin{tabular}{cccccccccc}
\toprule[1.0pt]
\multirow{2}{*}{\textbf{Method}} & \multicolumn{3}{c}{\textbf{\textsc{AToKe-SE}}} & \multicolumn{3}{c}{\textbf{\textsc{AToKe-ME}}} & \multicolumn{2}{c}{\textbf{\textsc{AToKe-EE}}}\\ \cmidrule{2-9} 
& CES    & CES-P      & HES     & CES    & CES-P      & HES  &CES &CES-P     \\ \hline

\multicolumn{9}{c}{\cellcolor{gray!25}\scshape GPT-J-6B \& 1-Edited} 
\\ \hline
FT$^*$        & 5.73     & 5.69   & 0.06       &1.11 &1.18 &0.03  &3.41  &2.91   \\
MEND$^*$    & 80.47     & 40.56   & 1.73       &71.83 &27.96 &0.40  &91.94  &62.48    \\ 
ROME$^*$    & \textbf{99.99}   & \textbf{97.01}    & 2.41    &\underline{98.85} &\underline{91.54} &0.44 &\underline{99.93}  &\underline{98.70}   \\ 
MEMIT$^*$    & 99.66  & 92.23    & 2.22      & 98.42  &91.06  &0.48 &99.92  &95.82   \\
FT$_{\text{METO}}$$^*$         & 2.8     & 2.62   & 3.38       &1.27 &1.2 &1.64  &-  &-   \\
MEND$_{\text{METO}}$$^*$     & 83.26     & 33.45   & 30.14       &70.52 &28.41 &28.65  &-  &-    \\ 
ROME$_{\text{METO}}$$^*$     & \underline{99.95}   & 93.78   & 20.25    &\textbf{99.93} &90.97 &23.22 &-  &-   \\ 
MEMIT$_{\text{METO}}$$^*$    & 86.4  & 85.32    & 30.31      & 92.73  &85.75  &36.2 &-  &-   \\
\hline
\multicolumn{9}{c}{\cellcolor{gray!25}\scshape GPT-J-6B \& All-Edited} \\ \hline
Mello  & 83.78   &81.55    &48.19    & 60.82    & 59.15   & 25.65 &\textbf{99.97}  &98.60 \\
PokeMQA   & 90.91   & 87.66    &\underline{62.49}     & 72.73    & 70.92   & \underline{40.99} &99.87  &98.62 \\
\OurMODEL{} (Ours)   & 97.95   & \underline{95.88}    & \textbf{97.25}    & 96.46    & \textbf{95.62}   & \textbf{96.43} &99.92 &\textbf{98.76} \\
\bottomrule[1.0pt]
\end{tabular}}
\caption{\textbf{Experiment results on \textsc{AToKe}.} METO is a plug-and-play method to strengthen the memory of historical knowledge. We mark the METO untested results as "-". \textsc{AToKe-ME} indicates each data includes two edits to perform continuous knowledge updating.}
\label{tab:exp2}
\vspace{-0.2in}
\end{table*}

\eat{As shown in Table ~\ref{tab:exp3}, Mello and PokeMQA experienced an average decrease of \warn{fix-me}\% and \warn{fix-me}\% respectively in M-Acc metric on \OurDataset{}-HK compared to \OurDataset{}. This is because Mello and PokeMQA match the embeddings of questions and edits for similarity, which easily leads to the loss of temporal information. Since \OurDataset{}-HK contains historical knowledge, the retrieval process of Mello and PokeMQA is prone to be influenced by historical knowledge, leading to incorrect answers.}



\begin{figure}[t]
\centering
\begin{tabular}{ccc}
\includegraphics[width=0.28\textwidth]{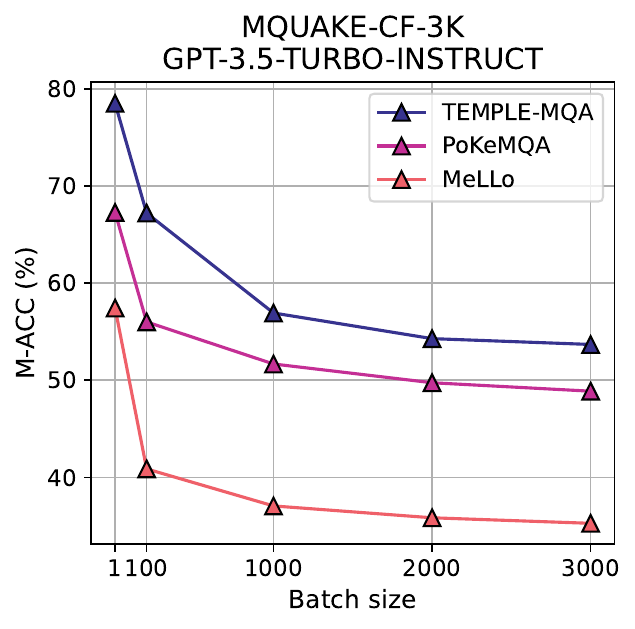} &
\includegraphics[width=0.28\textwidth]{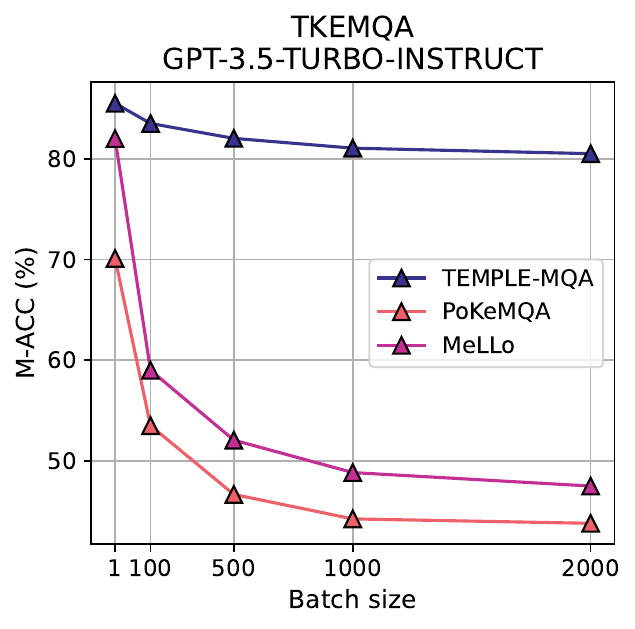} &
\includegraphics[width=0.28\textwidth]{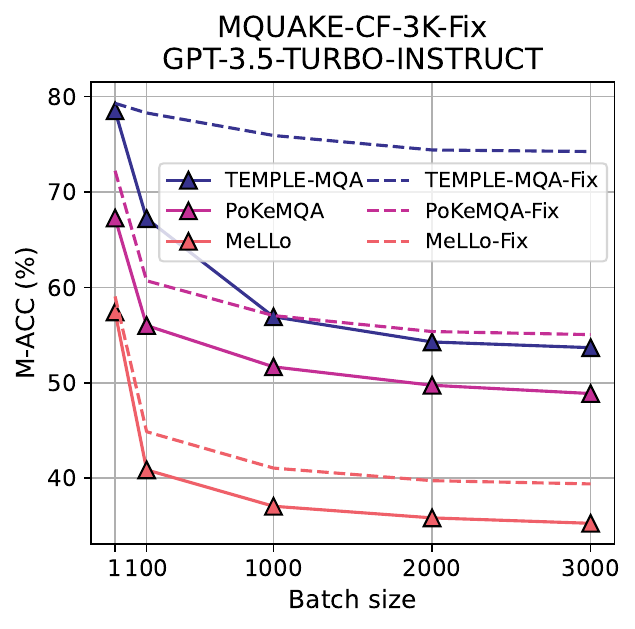} 
\end{tabular}
\caption{The line graph depicting the decrease in M-Acc with the increase in edit batch size on datasets \textsc{MQuAKE-CF-3K}, \OurDataset{} and \textsc{MQuAKE-CF-3K-Fix}.}
\label{fig:combined_fig1}
\vspace{-15pt}
\end{figure} 

\eat{
This indicate that PokeMQA overfits on \textsc{MQuAKE}. MeLLo uses self-checking to pass the retrieved edits to LLM to determine whether they are relevant, which is very effective when the retrieval accuracy is not high. For example, in the \OurDataset{}, the reason why PokeMQA performs poorly on HES is that the retrieved edit object is used as the answer, while Mello uses LLM to perform self-checking to determine whether the time matches. We discuss the use of self-checking in \OurMODEL{} more extensively in the appendix \warn{fix-me}.}

\eat{In conclusion, \OurMODEL{} can achieve good results under the dual challenges of MQA and temporal edit at the same time.}

\vspace{-0.05in}  
\subsection{Abalation Study}
\vspace{-0.05in} 
\label{sec:ablation}
\eat{
\vspace{-1.7ex}
In this section, we perform an in-depth analysis on the performance of
\OurMODEL{} under various settings, namely:
(i) Impact of inference path and retrieval via TAG,
(ii) MQA under Mass-Editing
(iii) Ablation analyses,
(iv) Error analyses.}

\eat{Mass editing is a key challenge for existing research on MQA under KE. This results in a decrease in retrieval accuracy, which has a high correlation with final performance. 
For instance, Zhong et al., highlighted that the accuracy of MeLLo~\citep{zhong2023mquake} declines by 37.2\% when the number 
of edits are increased beyond 3,000.}

We conduct an ablation study on different modules of the structural retrieval: without filtering sub-graphs according to the subject entity (i.e. - w/o Subject), without considering the similarity of relation in the Re-ranking step (i.e. - w/o Relation), without considering the similarity of the concept of object entity (i.e. - w/o Concept). As the inference path is the core component of our solution to MQA and cannot be ablated, so we only ablate different modules of the retrieval.

The results are shown in Table~\ref{abl-result} in Appendix. First, we investigate the role of filtering sub-graphs. We can see 
 without filtering out the sub-graph through the subject, \OurMODEL{} decreased by an average of 8.6\% and 7.8\% on M-Acc and H-Acc. This is the most impactful component among all removals.  Second, we can see both relation and concept can improve 
the performance of \OurMODEL{} in all the settings, indicating its necessity. Third, the impact of all eliminated components will become larger as the edit memory increases.

\section{Conclusion}
We presented a novel framework \OurMODEL{} for MQA under temporal knowledge editing. \OurMODEL{} first constructs a time-aware graph, then utilizes inference paths, structural retrieval, and joint reasoning, capable of enhancing MQA performance while distinguishing the temporal context. We also proposed a benchmark \OurDataset{}, which is the first benchmark specified for MQA with temporal scopes. Experiments on \OurDataset{} and the existing two benchmarks demonstrated that \OurMODEL{} outperforms previous methods.
\clearpage

\bibliography{colm2024_conference}
\bibliographystyle{colm2024_conference}
\newpage
\appendix

\section{Details of Prompts in \OurMODEL{}}\label{sec:prompt}

\definecolor{myTealLight}{rgb}{0.2, 0.8, 0.8} 
\begin{table*}[ht]
    \centering
    \small
    \begin{tabular}{l}
\toprule[1.0pt]
\color{red}\textbf{[Demo]}\color{black}\\
Unstructured fact: From 2009 to 2017, Obama was the president of the United States.\\

Structural form: <United States, the president is, person, 2009, 2017, Obama>\\

Unstructured fact: From 2019 to 2022, the head of government in kazakhstan is Askar Mamin.\\

Structural form: <Kazakhstan, head of government, person, 2019, 2022, Askar Mamin>\\
...Other demo ...\\
\color{teal}\textbf{[Instruction]}\color{black}\\
Please refer to the above demo and convert the following edit into the structural form: \\<subject, relation, type of object, start time, end time, object>\\
\color{brown}\textbf{[Task]}\color{black}\\
<Fact>\\
\bottomrule[1.0pt]

\end{tabular}
    \caption{\textbf{The prompts $P_{convert}$ to convert unstructured knowledge used in Equation (\ref{eq:convert1})}. \color{red}\textbf{[Demo] }\color{black} include several  high-quality demonstrations handwritten by humans. <Fact> indicates the unstructured fact $f_x$. More examples are shown in Table~\ref{tab:convert_edit}.}
    \label{tab:prompt_for_convert_edit}
\end{table*}

\begin{table*}[ht]
    \centering
    \small
    \begin{tabular}{l}
\toprule[1.0pt]
\color{red}\textbf{[Demo]}\color{black}\\
Question: Who was the owner of Tom's company from 2018 to 2020?\\
Inference path: <Tom, company is, company>, <company, owner is, people>\\
Time: 2018, 2020\\
Question: What is the religion of the head of government of Israel after 2022?\\
Inference path: <Israel, head of government, person>, <person, religion is, religion>\\
Time: 2022, None\\
...Other demo...\\
\color{teal}\textbf{[Instruction]}\color{black}\\
Please refer to the above demo and generate a valid inference path for the following question:\\
\color{brown}\textbf{[Task]}\color{black}\\
<Question>\\
\bottomrule[1.0pt]
\end{tabular}
    \caption{\textbf{The prompt $P_{plan}$ to extract inference path used in Equation (\ref{eq:Stage1})}, where <Question> indicates the question $Q$.}
    \label{tab:prompt_for_extract_inference_path}
\end{table*}

\begin{table*}[ht]
    \centering
    \small
    \begin{tabular}{l}
\toprule[1.0pt]
\color{red}\textbf{[Demo]}\color{black}\\
Inference step: <United Kingdom, head of government, person>\\
Time: from 2019 to 2022\\
Answer: Boris Johnson\\

Inference step: <Christian Wulff, spouse, person>\\
Time: after 2023\\
Answer: Bettina Wulff\\

...Other demo...\\
\color{teal}\textbf{[Instruction]}\color{black}\\
Please generate a valid answer like the above example for the following task:\\
\color{brown}\textbf{[Task]}\color{black}\\
<Inference step>\\
<Time>\\
\bottomrule[1.0pt]
\end{tabular}
    \caption{\textbf{The prompt $P_{query}$ for querying LLM to generate answer used in Equation (\ref{eq:retrieveanswer})}, where <Inference step> indicates specific inference step in the inference path, <Time> indicates temporal scope.}
    \label{tab:prompt_for_gen_answer}
\end{table*}

\section{Experimental Details}
\label{Append:exp-detail}

\subsection{Dataset}
\label{appendix:Dataset}
We provide a detailed description of the evaluation datasets as follows.

\textbf{(a) \textsc{MQuAKE} \citep{zhong2023mquake}.} \textsc{MQuAKE} includes \textsc{MQuAKE-CF-3K} based on counterfactual editing and \textsc{MQuAKE-T} based on real-world changes. These datasets encompass k-hop questions ($k\in \{2,3,4\}$), each associated with one or more edits. Statistics are shown in Table ~\ref{tab:dataset}. 

\textbf{(b) \textsc{AToKe} \citep{Yin2023HistoryMT}.} 
\textsc{AToKe} is the first temporal knowledge editing dataset containing a series of world knowledge with timestamps, regarded as a series of knowledge updates. \textsc{AToKe} contains
three editing types: single edit (SE), multiple edits (ME), and extending edit (EE), corresponding to three datasets \textsc{AToKe-SE}, \textsc{AToKe-ME} and \textsc{AToKe-EE}. Statistics of these three datasets are shown in Table~\ref{tab:dataset3}.
\begin{itemize}
    \item \textbf{\textsc{AToKe-SE}.} Each data of this dataset includes one edit for knowledge updating and a historical question and current question corresponding to the time scope before and after the modification. 
    \item \textbf{\textsc{AToKe-ME}.}  Each data of this dataset includes two edits to perform continuous knowledge updating, two historical questions, and one current question.   
    \item \textbf{\textsc{AToKe-EE}.} Each data of this datasets include one edit to modify the time scope of knowledge (not updating the object), and only one current question. 
\end{itemize}

It is worth explaining that the above three datasets are all aimed at single-hop questions only.

\textbf{(c) \OurDataset{}.} \OurDataset{} is our newly curated benchmark dataset. It is a blend of carefully crafted multi-hop questions with explicit temporal scopes.
This data set is designed to evaluate the ability of KE methods rigorously. \textbf{\OurDataset{}-HK} provides additional knowledge corresponding to two historical questions on the basis of \OurDataset{}, which is used to explore whether the memory-based method will confuse the temporal context. See Section \ref{sec:ourdata} for a detailed introduction, construction process, and statement temples of \OurDataset{}.

\begin{table*}
\centering
\resizebox{0.55\linewidth}{!}{
\begin{tabular}{llllll}
\toprule[1.0pt]
\textbf{Datasets} & \textbf{\#Edits} & \textbf{2-hop} & \textbf{3-hop} & \textbf{4-hop} & \textbf{Total}\\
\midrule
\multirow{5}{*}{\textbf{\textsc{MQuAKE-CF-3K}}} &1  &513  &356  &224 &1093 \\
&2  &487  &334  &246 &1067 \\  
&3  &-  &310  &262 &572 \\
&4  &-  &-  &268 &268 \\
&All  &1000  &1000  &1000 &3000 \\ 
\midrule
\textbf{\textsc{MQuAKE-T}} &1  &1421  &445  &2 &1868 \\
\bottomrule[1.0pt]
\end{tabular}}
\caption{\textbf{Statistics of the \textsc{MQuAke} dataset}. }
\label{tab:dataset}
\end{table*}

\begin{table*}
\centering
\resizebox{0.55\linewidth}{!}{
\begin{tabular}{cccc}
\toprule[1.0pt]
\textbf{Datasets} & \textbf{\textsc{AToKe-SE}} & \textbf{\textsc{AToKe-ME}} &  \textbf{\textsc{AToKe-EE}} \\
\midrule
\textbf{Size} &8819 &8820 &8819\\
\bottomrule[1.0pt]
\end{tabular}}
\caption{\textbf{Statistics of the \textsc{AToKe} dataset}. }
\label{tab:dataset3}
\end{table*}

\begin{table*}
\centering
\resizebox{0.65\linewidth}{!}{
\begin{tabular}{cccc|ccccc}
\toprule[1.0pt]
\textbf{Datasets} & \textbf{\#UK}& \textbf{\#CK}& \textbf{\#HK}  & \textbf{1-hop}& \textbf{2-hop} & \textbf{3-hop} & \textbf{4-hop} & \textbf{Total}\\
\midrule
\multirow{3}{*}{\textbf{\OurDataset{}}} &1  &0  &0  &500 &334 &- &- &834\\
&1  &1  &0 &- &166 &500 &500 &1166\\ 
\cmidrule{2-9}
&\multicolumn{3}{c|}{\textbf{All}} &500 &500 &500 &500 &2000\\
\midrule
\multirow{3}{*}{\textbf{\OurDataset{}-HK}} &1  &0  &2  &500 &334 &- &- &834\\
&1  &1  &2 &- &166 &500 &500 &1166\\  
\cmidrule{2-9}
&\multicolumn{3}{c|}{\textbf{All}} &500 &500 &500 &500 &2000\\
\bottomrule[1.0pt]
\end{tabular}}
\caption{\textbf{Statistics of \OurDataset{} and \OurDataset{}-HK}. \textbf{\#UK} represents the number of updated knowledge, \textbf{\#CK} represents the number of corrective knowledge and \textbf{\#HK} represents the number of historical knowledge. \OurDataset{}-HK include two questions asking about historical knowledge which \OurDataset{} does not include.}
\label{tab:dataset2}
\end{table*}
\subsection{Baseline Models}
\label{Appendix:baseline}
\textbf{(a) Parameter-based.} The parameter-based baselines include: 
(i) \textbf{Fine-tuning (FT)}~\citep{Zhu2020ModifyingMI} that 
performs a gradient-based update on the model parameters
to incorporate new knowledge. 
(ii) \textbf{ROME}~\citep{meng2022locating} first locates factual 
knowledge at a specific layer of the transformer architecture and 
then updates the feed-forward network of this layer to insert new 
knowledge. 
(iii) \textbf{MEMIT}~\citep{meng2022mass} extends ROME to allow 
modifying a range of feed-forward network layers for a large amount 
of knowledge. 
(iv) \textbf{MEND} \citep{Mitchell2021FastME} learns a hyper network to produce
weight updates by decomposing
the gradient of standard fine-tuning into a low-rank form.
(v) \textbf{METO}~\citep{Yin2023HistoryMT} extends the parameter-based 
methods by adding the time information as an optimization objective. 
\eat{also allowing multiple edits on both historical and current knowledge.}

\noindent \textbf{(b) Memory-based.} The memory-based baselines include: 
(i) \textbf{MeLLo}~\citep{zhong2023mquake} that uses the plan-and-solve 
paradigm. \eat{When solving a sub-problem, MeLLo first asks the model to 
generate a candidate answer and then give the retrieved edits to 
the model to determine whether it is relevant.}
(ii) \textbf{PokeMQA}~\citep{gu2023pokemqa} extends MeLLo by
adopting a two-stage retrieval process to decouple the 
question decomposition and knowledge editing. 
\subsection{Evaluation Metrics}
\label{Append:evalm}

Details about the evaluation metrics and their mathematical formulation are provided as follows: 

(i) \textbf{Multi-hop Accuracy (M-Acc)}, which is used 
to measure the accuracy of the language models on multi-hop questions.
For M-Acc, we use the same settings as ~\citet{zhong2023mquake}. The calculation formula for M-Acc is as follows:
\begin{equation}
\mathbbm{1}\left[\bigvee_{q \in \mathcal{Q}} [f^*(q) = a^*]\right].
\end{equation}
Where $f^*(\cdot)$ represents the edited model, $\mathcal{Q}$ and $a^*$ represent the multi-hop questions and the edited answer for each case, respectively.

(ii) \textbf{Hop-wise Accuracy (H-Acc)}, which is used to check the correctness 
of the intermediate reasoning path for MQA. For H-Acc, we follow the same 
settings proposed by ~\citet{gu2023pokemqa}. Given edited chain of facts $\mathcal{C}^*$, H-Acc is defined as
\begin{equation}
\mathbbm{1}\left[\bigwedge_{(s, r, o^*) \in \mathcal{C}^*}[f^*(s,r)=o^*]\right].
\end{equation}

(iii) \textbf{Historical Explicit time Question Score (HES)}, which is the 
accuracy of explicit temporal questions about historical knowledge~\citep{Yin2023HistoryMT}.

(iv) \textbf{Current Explicit time Question Score (CES)}, which measures 
the accuracy of explicit temporal questions about current knowledge~\citep{Yin2023HistoryMT}.

(v) \textbf{Current Explicit time
Paraphrase Question Score (CES-P)}, which is 
the accuracy of explicit temporal questions of semantic paraphrasing asked about 
current knowledge~\citep{Yin2023HistoryMT}.

HES, CES, and CES-P require the model to be able to recall knowledge at a specific temporal scope,  calculated according to the following formula:
\begin{equation}
\mathbbm{1}\left[f^*(s,r,\tau_s,\tau _e)=o\right].
\end{equation}


\eat{
\begin{equation}
   sim(v_1,v_2)=\frac{v_1 \cdot v_2}{\lVert v_1 \rVert \lVert v_2 \rVert}
    \label{eq:SIM}
\end{equation}} 

\section{More Experiment Results}\label{sec:more}

\begin{table*}[!htb]
\centering
\resizebox{0.9\linewidth}{!}{
\begin{tabular}{ccccccccccc}
\toprule[1.0pt]
\multirow{3}{*}{\textbf{Method}}
  &\multicolumn{6}{c}{\textbf{\textsc{MQuAKE-CF-3K}}} &\multicolumn{4}{c}{\textbf{\textsc{MQuAKE-T}}} \\ \cmidrule{2-11} 
  &\multicolumn{2}{c}{1-edited} &\multicolumn{2}{c}{100-edited} &\multicolumn{2}{c}{All-edited} &\multicolumn{2}{c}{1-edited} &\multicolumn{2}{c}{All-edited}\\ \cmidrule{2-11} 
&M-Acc &H-Acc &M-Acc &H-Acc &M-Acc &H-Acc &M-Acc &H-Acc &M-Acc &H-Acc\\
\midrule
- w/o Subject   &77.96 &62.85 &62.06 &51.16 &44.09 &34.23 &89.56 &80.77 &73.97 &66.56   \\
- w/o Relation   &77.78 &62.67 &65.15 &52.88 &49.53 &36.51 &89.72 &80.80 &77.13 &68.49 \\
- w/o Concept  &\underline{78.08} &\underline{63.02} &\underline{65.70} &\underline{53.29} &\underline{50.10} &\underline{37.55} &\underline{90.01} &\underline{81.35} &\underline{78.67} &\underline{70.20}  \\
\midrule
\OurMODEL{}    & \textbf{78.11} & \textbf{63.45}  & \textbf{67.21} &\textbf{55.33}   & \textbf{53.68}  &\textbf{40.05} & \textbf{90.57} &\textbf{81.90}  & \textbf{82.26} &\textbf{74.33}  \\
\bottomrule[1.0pt]
\end{tabular}}
\caption{Ablation study result of \OurMODEL{}.}
\label{abl-result}
\end{table*}

\begin{table*}[!htb]
\centering
\resizebox{0.75\linewidth}{!}{
\begin{tabular}{ccccccc}
\toprule[1.0pt]
\multirow{2}{*}{\textbf{Method}} & \multicolumn{2}{c}{\textbf{2-hop}} & \multicolumn{2}{c}{\textbf{3-hop}} & \multicolumn{2}{c}{\textbf{4-hop}}\\ \cmidrule{2-7} 
& Cost(\$)    & Time(s)      & Cost(\$)     & Time(s)     & Cost(\$)      & Time(s)        \\ 
\midrule
\textbf{Mello} &0.73 &8.98 &0.94 &12.13 &1.61 &16.41\\

\textbf{PokeMQA} &0.93 &10.33 &1.09 &11.98 &1.36 &18.85\\ 

\textbf{\OurMODEL{}} &0.15 &2.50 &0.25 &5.31 &0.27 &6.98\\
\bottomrule[1.0pt]
\end{tabular}}
\caption{Average expense and time cost comparison of methods on the \textsc{MQuAKE-CF-3K} dataset for solving per multi-hop problem in the same experimental environment.}
\label{tab:time_and_cost}
\end{table*}

\begin{table*}[!htb]
\centering
\resizebox{0.9\linewidth}{!}{
\begin{tabular}{cccccccccc}
\toprule[1.0pt]
\multirow{2}{*}{\textbf{Method}} & \multicolumn{3}{c}{\textbf{1-edited}} & \multicolumn{3}{c}{\textbf{100-edited}} & \multicolumn{3}{c}{\textbf{All-edited}}\\ \cmidrule{2-10} 
& R-Acc   & M-Acc      & H-Acc     & R-Acc    & M-Acc      & H-Acc     & R-Acc    & M-Acc      & H-Acc  \\
\midrule
\textbf{Mello} &91.6 &57.43 &28.8 &83.1 &40.87 &28.13 &74.5 &35.27 &25.3\\

\textbf{PokeMQA} &95.2 &67.27 &56.37 &89.5 &56.00 &49.63 &81.5 &48.87 &39.77\\

\textbf{\OurMODEL{}} &99.5 &78.11 &63.45 &93.2 &67.21 &55.33 &87.8 &53.68 &40.05\\
\bottomrule[1.0pt]
\end{tabular}}
\caption{ Retrieval accuracy of different methods on the \textsc{MQuAKE-CF-3K} dataset. R-Acc refers to the accuracy of retrieving edited facts from memory.}
\label{tab:retrieving_acc}
\end{table*}

\textbf{Data Conflict in \textsc{MQuAKE}.} In the analysis for MQA under Temporal KE, we observed that \OurMODEL{} on \OurDataset{}
experiences only a slight performance decrease when the edit batch size increases, while its decrease is faster for \textsc{MQuAKE}. 
This is because \textsc{MQuAKE-CF-3K}  has unpassable data
in the mass-edit setting, i.e., we can observe data conflicts in \textsc{MQuAKE}. These conflicts arise from inconsistent counterfactual edits between two cases containing the same knowledge (i.e., one is edited, but another is not edited). Under mass editing settings, the inference process of one case may be influenced by the editing of another case, leading to incorrect answers. Table~\ref{tab:data_conflict} provides an example of data conflict in \textsc{MQuAKE}.

\textbf{Cost of Inference Path.} 
Here, we aim to compare the expense and time cost of our inference path to those of previous methods. 
According to Table~\ref{tab:time_and_cost}, we can easily see that, for \textsc{MQuAKE-CF-3K}, \OurMODEL{} averagely saves 78.7\% and 80.4\% of the expense compared to Mello and PokeMQA, respectively, and reduces the inference time by 62.9\% and 64.8\%. The reason is that \OurMODEL{} only needs to call LLMs once to generate the inference path, while Mello and PokeMQA need to call LLMs multiple times to generate sub-questions. Thus, we claim \OurMODEL{} offers a faster and more cost-effective solution.

\textbf{Accuracy of Structured Retrieval.} To achieve reliable KE, apart from the correct decomposition of sub-problems, another important factor is whether related edits can be found correctly.
Here, we conduct experiments to compare the accuracy of the retrieval process for different methods. According to Table~\ref{tab:retrieving_acc}, \OurMODEL{} achieve higher retrieval accuracy compared to Mello and PokeMQA by 17.9\% and 7.7\% in the all-edited setting, respectively. Thus, our structured retrieval can efficiently find structured edits.

\begin{figure}[t]
\centering
\begin{subfigure}{0.49\textwidth}
  \centering
  \includegraphics[width=\linewidth]{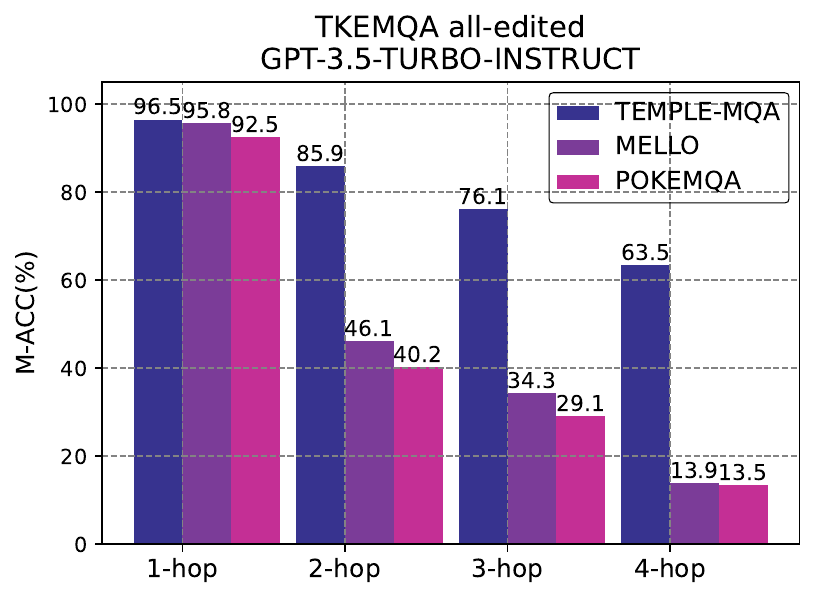}
\end{subfigure}%
\begin{subfigure}{0.42\textwidth}
  \centering
  \includegraphics[width=\linewidth]{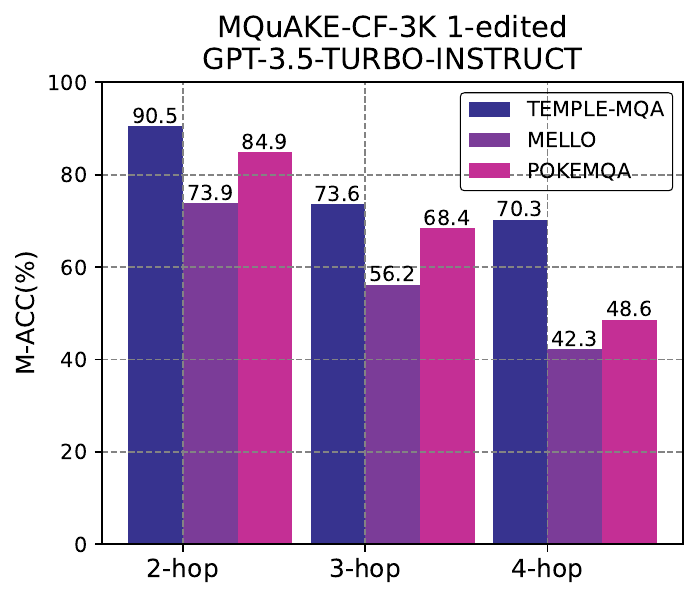}
\end{subfigure}
\caption{A bar graph comparing the M-Acc of three methods under two settings: All-edited for \OurMODEL{} and 1-edited for \textsc{MQuAKE-CF-3K}.}
\label{fig:combined_fig2}
\end{figure}

\textbf{\OurMODEL{} under Different Hop Numbers.}
Here, we aim to study the performance of different models with different hop numbers. As shown in Figure ~\ref{fig:combined_fig2}, \OurMODEL{} can still maintain a high M-Acc as the number of hops increases, while Mello and PokeMQA experience an obvious decline in performance with an increase in hop number. This is because Mello and PokeMQA couple the tasks of querying the LLM to answer sub-questions and generate sub-questions together. As the hop number increases, the LLM is burdened with complex tasks, causing a significant decrease in M-Acc. In contrast, \OurMODEL{} decouples the generation of inference paths, enabling it to maintain high performance as the hop number increases.

\begin{table*}[ht]
    \centering
    \small
    \scalebox{0.9}{
    \begin{tabular}{cl}
\toprule
\multicolumn{2}{c}{\scshape  \textbf{Data 998 in \textsc{MQuAKE-CF-3K}}}\\
\midrule
\textbf{Questions} & What continent does Gareth David-Lloyd hold citizenship in?\\
&In which continent was the actor Gareth David-Lloyd a citizen of?\\
&Which continent does the citizenship of Gareth David-Lloyd belong to?\\
\midrule
\textbf{Edits} & (Gareth David-Lloyd, country of citizenship, Nigeria)\\
&\sethlcolor{c1}\hl{(Nigeria, continent, North America)}\\
\toprule
    \multicolumn{2}{c}{\scshape  \textbf{Data 706 in \textsc{MQuAKE-CF-3K}}}\\
\midrule
\textbf{Questions} & What continent does the country Peter Green belong to citizenship to belong to?\\
&Which continent is the country of citizenship of Peter Green?\\
&On which continent is the country where Peter Green holds citizenship located?\\
\midrule
\textbf{Edits} & (Peter Green, country of citizenship, Nigeria)\\
\toprule
\multicolumn{2}{c}{\scshape \textbf{Original inference procedure in Data 706}}\\
\midrule
\textbf{Inference procedure} & (Peter Green, country of citizenship, Nigeria)\\
&\sethlcolor{c2}\hl{(Nigeria, continent, Africa)}\\
\midrule
\textbf{Original Answer} & Africa\\
\toprule
\multicolumn{2}{c}{\scshape  \textbf{Disturbed inference procedure in Data 706}}\\
\midrule
\textbf{Inference procedure} & (Peter Green, country of citizenship, Nigeria)\\
&\sethlcolor{c1}\hl{(Nigeria, continent, North America)}\\
\midrule
\textbf{Wrong Answer} & North America\\
\bottomrule
\end{tabular}}
    \caption{\textbf{Data Conflict in \textsc{MQuAKE-CF-3K}.}  Under the setting of mass editing, a step (highlighted in \sethlcolor{c2}\hl{blue}) in the original inference procedure in Date 706 is changed due to retrieving an edit in Data 998 (highlighted in \sethlcolor{c1}\hl{red}), resulting in a wrong answer.}
    \label{tab:data_conflict}
\end{table*}

\section{Temporal Knowledge Editing for Multi-hop Question Answering Benchmark: \OurDataset{}} \label{sec:ourdata}
\begin{table*}[ht]
    \centering
    \small
    \resizebox{0.9\linewidth}{!}{
    \begin{tabular}{cl}
\toprule
\multicolumn{2}{c}{\scshape \textbf{Questions}}\\
\midrule
\textbf{Current question} & In which continent is the country where the head of government\\ &of Taiwan has citizenship located after 2023?\\
\midrule
\textbf{Historical questions} & Who is the head of government in Taiwan from 2019 to 2023?\\ &Who is the head of government in Taiwan from 2017 to 2019?\\
\toprule
\multicolumn{2}{c}{ \scshape \textbf{Knowledge store in Memory}}\\
\midrule
\textbf{Corrective knowledge} & \sethlcolor{c2}\hl{(Chen Chien-jen, country of citizenship, Taiwan $\to$ Algeria)}\\
\midrule
\textbf{Updated knowledge}
&\sethlcolor{c1}\hl{(Taiwan, head of government, Chen Chien-jen, 2023, N/A)}\\
\midrule
\textbf{Historical knowledge}
&(Taiwan, head of government, Hope Su, 2019, 2023)\\
&(Taiwan, head of government, Lai Ching-te, 2017, 2019)\\
\toprule
\multicolumn{2}{c}{ \scshape  \textbf{Procedure to answer Current question}}\\
\midrule
\textbf{Inference procedure} & \sethlcolor{c1}\hl{(Taiwan, head of government, Chen Chien-jen)}\\
&\sethlcolor{c2}\hl{(Chen Chien-jen, country of citizenship, Algeria)}\\
&(Algeria, continent, Africa)\\
\toprule
\multicolumn{2}{c}{\scshape \textbf{Answers}}\\
\midrule
\textbf{Current answer} & Africa \\
\midrule
\textbf{Historical answers} & Hope Su \\ &Lai Ching-te\\
\bottomrule

\end{tabular}}
    \caption{\textbf{An instance in \OurDataset{}-HK.} Updated knowledge refers to the updating of outdated knowledge and does not deny the correctness of historical knowledge. Corrective knowledge refers to the subversion of a model's past knowledge used to simulate the model's modification of erroneous knowledge. The updated knowledge and corrective knowledge required in the inference procedure are highlighted in \sethlcolor{c1}\hl{red} and \sethlcolor{c2}\hl{blue}, respectively.}
    \label{tab:q_examples}
\end{table*}
\subsection{Introduction to \OurDataset{}}
Our benchmark \OurDataset{} is created based on Wikidata \footnote{\url{https://www.wikidata.org/wiki/Wikidata:Main_Page}}, a knowledge base consisting of fact triples associated with millions of entities. \OurDataset{} can be used to evaluate whether the knowledge editing method can update new knowledge without forgetting past knowledge. It contains simple single questions and multi-hop questions, and its rich settings enable it to comprehensively evaluate knowledge editing methods.

As shown in Table~\ref{tab:dataset2}, \OurDataset{} contains 2,000 data where each one includes a k-hop $(k\in \{1,2,3,4\})$ questions. As GPT-3.5 does not have the knowledge after 2022, we use the knowledge after 2022 as an edit to \textit{update knowledge}. Thus, we refer to each multi-hop question as a current question because it has a time prompt after 2022. Besides the updated knowledge, we also introduce the \textit{corrective knowledge}, which is used to modify the error historical knowledge. This knowledge does not need temporal scope and is counterfactual. 

To evaluate whether the editing method may cause model forget the historical knowledge (i.e. knowledge of model before the editing), each data of \OurDataset{} include two historical question with the time prompt before 2022. Based on \OurDataset{}, we add the knowledge corresponding to the historical question into the edit memory to evaluate whether the retrieval can effectively distinguish different knowledge in the same time chain, in which we refer to as \textsc{TKeMqa-HK}.

\eat{
Each case in \OurDataset{} contains a k-hop chain of facts containing one updated knowledge and at most one counterfactual edit, as well as question, answer, and two historical knowledge. The specific format for each data is shown in Table ~\ref{tab:q_examples}.}

We use M-Acc and HES as two metrics in \OurDataset{}. M-Acc is the accuracy of the language models on current multi-hop questions, evaluating the model's ability to update knowledge and perform multi-hop reasoning successfully. HES is the accuracy of answering historical questions and assessing the model's ability to recall historical knowledge.

\subsection{\OurDataset{} Data Curation}

\textbf{Collect relation template.} We first manually collect multiple relations containing time information from Wikidata (Appendix \ref{sec:cloze}). The relation serves as a fact template to obtain objects from Wikidata using SPARQL when given a subject.     

\textbf{Generate relation path template.} To construct multi-hop questions, we arrange and combine all the collected relations into relation path \citep{Luo2023ReasoningOG} with different lengths $k\in \{1,2,3,4\}$. The relation path is used as a chain of facts template to generate the inference procedure for multi-hop questions when given the start point entity.

\textbf{Generate question template for relation path template.}
Based on the chain of facts template to generate data, we can get the relation and object of each hop. In order to generate multi-hop questions in the form of natural language, we use GPT-4 to generate multi-hop questions template for each relation path template. Different from previous work (\citet{zhong2023mquake}), this method produces data with higher quality questions due to the powerful capabilities of GPT-4. Using the relation path template also reduces the huge cost of GPT-4 (about ten times that of GPT-3.5). 
 
\textbf{Generate data based on the template.} After creating the template mentioned above, we randomly select the starting point of the relation path to generate multi-hop questions. We set the rule that each multi-hop question has exactly one knowledge after 2022, which mimics the knowledge update.

\subsection{Question/Cloze Statement Templates used in \OurDataset{}}\label{sec:cloze}
Following previous work, we use question templates to filter the facts that cannot be recalled and cloze-style statement templates to convert an edited fact into a natural language form statement. Table ~\ref{tab:templates} shows question templates and cloze-style statement templates employed in \OurDataset{}.
\begin{table*}[ht]
    \resizebox{0.95\linewidth}{!}{
    \begin{tabular}{l|l|l}
    \toprule 
     Relation & Question template & Cloze-style statement template \\
     \midrule
P35 & Who is the head of state in [S]? & The head of state in [S] is\\
P6 & Who is the head of government in [S]? & The head of government in [S] is\\
P488 & Who is the chairperson of [S]? & The chairperson of [S] is\\
P169 & Who is the chief executive officer of [S]? & The chief executive officer of [S] is\\
P54 & Which sports team is [S] affiliated with? & [S] is affiliated with the sports team of\\
P286 & Who is the head coach of [S]? & The head coach of [S] is\\
P551 & Where does [S] live? & [S] live in\\
P102 & Which political party is [S] affiliated with? & [S] is affiliated with the political party of\\
P26 & Who is [S]'s spouse? & [S]'s spouse is\\
P38 & What is the currency of [S]? & The currency of [S] is\\
P108 & Which organization is [S] an employee of? & [S] is an employee of\\
P69 & Which university is [S] educated at? & [S] is educated at\\
P937 & Where is [S]'s workplace? & The work location of [S] is\\
P36 & What is the capital of [S]? & The capital of [S] is\\
P159 & Where is the headquarters of [S] located? & The headquarters of [S] is located in\\
P27 & What is the country of citizenship of [S]? & [S] is a citizen of\\
P140 & Which religion is [S]affiliated with? & [S] is affiliated with the religion of\\
P30 & Which continent is [S] located in? & [S] is located in the continent of\\
P37 & What is the official language of [S]? & The official language of [S] is\\
P17 & Which country is [S] located in? & [S] is located in the country of\\

\bottomrule
    \end{tabular}}
    \caption{Question templates and cloze-style statement templates employed in \OurDataset{}. "[S]" denotes a placeholder for the subject entity within the fact. The question templates filter the facts that cannot be recalled. The cloze-style statement templates convert an edited fact into a natural language form statement. }
    \label{tab:templates}
\end{table*}

\section{SPARQL Protocol and RDF Query Language}\label{sec:SPARQL}
SPARQL is able to retrieve and manipulate data stored in Resource Description Framework (RDF) format, which can store graph information. Wikidata Query Service (\href{https://query.wikidata.org/}{WDQS}) is an online tool for querying the Wikidata database. It allows users to retrieve and analyze structured data in Wikidata using the SPARQL query language. We use SPARQL to access WDQS to obtain the alias of the subject entity and create our dataset \OurDataset{} (see the table \ref{tab:sparql}). 

\begin{table*}[ht]
    \centering
    \small
    \resizebox{0.8\linewidth}{!}{
    \begin{tabular}{cl}
\toprule
\multicolumn{2}{c}{\textbf{Using SPARQL to extract aliases}}\\
\midrule
\multirow{4}{*}{\textbf{SPARQL}} &SELECT ?alias WHERE \{\\
&\quad wd:<QID> skos:altLabel ?alias.\\
&\quad FILTER(LANG(?alias) = "en").\\
&\} \\
\midrule
\multirow{3}{*}{\textbf{Description}} &This SPARQL query retrieves English alias for a specific\\& entity. "<QID>" denotes a placeholder for the entity id in\\& Wikidata.\\
\toprule
\multicolumn{2}{c}{\textbf{Using SPARQL to extract facts}}\\
\midrule
\multirow{4}{*}{\textbf{SPARQL}} &SELECT ?x ?Label
WHERE \{\\
&\quad wd:<QID> wdt:<PID> ?x.\\
&\quad ?x rdfs:label ?Label.\\
&\quad FILTER(LANG(?Label) = "en")\\
&\}\\
\midrule
\multirow{3}{*}{\textbf{Description}} &This SPARQL is used to retrieves corresponding object entity\\& according to the subject entity id (represented by "<QID>") \\&and relation id (represented by "<PID>").\\
\toprule
\multicolumn{2}{c}{\textbf{Using SPARQL to extract facts with temporal scope}}\\
\midrule
\multirow{9}{*}{\textbf{SPARQL}} &SELECT ?x ?Label ?start\_time ?end\_time WHERE \{\\
&\quad wd:<QID> p:<PID> ?statement.\\
&\quad ?statement ps:<PID> ?x.\\
&\quad ?x rdfs:label ?Label.\\
&\quad OPTIONAL\{ ?statement pq:P580 ?start\_time.\}\\
&\quad OPTIONAL\{ ?statement pq:P582 ?end\_time.\}\\ 
&\quad FILTER(LANG(?Label) = "en").\\  
&\}\\
&ORDER BY DESC (?start\_time)\\
\midrule
\multirow{3}{*}{\textbf{Description}} &This SPARQL query is used to extract object entities and their \\&corresponding time scopes based on subject entity id (denoted by \\&"<QID>") and relation id (denoted by "<PID>"). \\

\bottomrule

\end{tabular}}
    \caption{ \textbf{The SPARQL we used to create \OurDataset{}.}}
    \label{tab:sparql}
\end{table*}

\begin{table*}[ht]
    \resizebox{1.0 \linewidth}{!}{
    \begin{tabular}{l|l}
    \toprule 
     Unstructured Form & Structured Form\\
     \midrule
From 1993 to 1999, Donald Trump’s spouse is Marla Maples.&(Donald Trump, spouse, person, 1993, 1999, Marla Maples)\\
In 2020, the head coach of AC Horsens is Jonas Dal. &(AC Horsens, head coach, person, 2020, 2020, Jonas Dal)\\
After 2023, the head of government in Slovakia is Robert Fico. &(Slovakia, head of government, person, 2023, N/A, Robert Fico)\\
Before 1936, the head of state in United Kingdom is George V. &(United Kingdom, head of state, person, N/A, 1936, George V)\\
The capital of United Kingdom is London. &(United Kingdom, capital, city, N/A, N/A, Londo)\\
\bottomrule
    \end{tabular}}
    \caption{Examples of transforming knowledge from unstructured form into structured form. The N/A indicate unknown for start time or end time.. }
    \label{tab:convert_edit}
\end{table*}
\section{Limitation}
Although Large Language Models (LLMs) have garnered widespread attention for their remarkable capacity for knowledge comprehension \citep{yang2024human,yang2024moral}, enabling tailored solutions across various applications, they also face critical issues such as privacy concerns \citep{hu2023differentially}, and explainability \citep{hu2023seat,lai2023faithful}. LLM applications typically involve data containing sensitive information, necessitating effective solutions to safeguard privacy \citep{xu2023llm}. One promising approach to address this challenge is the design of Differentially Private (DP) algorithms \citep{dwork2006calibrating}. DP offers provable protection against identification and is resilient to arbitrary auxiliary information that may be available to attackers. While there have been numerous studies on DP machine learning \citep{hu2022high,wang2020differentially,wang2021estimating,su2022faster,hu2023privacy} and DP deep learning \citep{xiang2024does,xiang2023practical}, most of these efforts have primarily focused on either continuous tabular data or image data. Unfortunately, less attention has been given to adapting variants of DP algorithms to the context of Natural Language Processing (NLP) and the text domain. Addressing this gap is crucial as text data presents its own unique challenges and characteristics that necessitate specialized privacy-preserving techniques. By developing and refining DP algorithms tailored to NLP tasks, we can enhance the privacy protections of LLMs and enable their responsible and ethical deployment across various domains. We will leave it for future work.

\end{document}